\documentclass[letterpaper, 10 pt, conference]{ieeeconf}  

\IEEEoverridecommandlockouts                              

\usepackage{graphics} 
\usepackage{epsfig} 
\usepackage{mathptmx} 
\usepackage{times} 
\usepackage{amsmath} 
\usepackage{amssymb}  
\usepackage{hyperref}
\usepackage{url}
\usepackage[noend]{algorithmic}
\usepackage[noend,ruled,vlined,linesnumbered,commentsnumbered]{algorithm2e}

\title{\LARGE \bf A toolbox for neuromorphic perception in robotics}

\author{Julien Dupeyroux, Stein Stroobants and Guido C.H.E. de Croon
\thanks{*This work has received funding from the ECSEL Joint Undertaking (JU) under grant agreement No. 826610. The JU receives support from the European Union's Horizon 2020 research and innovation program and Spain, Austria, Belgium, Czech Republic, France, Italy, Latvia, Netherlands. It has also received funding from the European Office of Aerospace Research and Development (EOARD).}
\thanks{All authors are with the Micro Air Vehicle Lab, Faculty of Aerospace
Engineering, Delft University of Technology, The Netherlands. Contact: 
        {\tt\small j.j.g.dupeyroux@tudelft.nl}}%
}

\begin{document}

\maketitle
\thispagestyle{empty}
\pagestyle{empty}

\begin{abstract}
The third generation of artificial intelligence (AI) introduced by neuromorphic computing is revolutionizing the way robots and autonomous systems can sense the world, process the information, and interact with their environment. The promises of high flexibility, energy efficiency, and robustness of neuromorphic systems is widely supported by software tools for simulating spiking neural networks, and hardware integration (neuromorphic processors). Yet, while efforts have been made on neuromorphic vision (event-based cameras), it is worth noting that most of the sensors available for robotics remain inherently incompatible with neuromorphic computing, where information is encoded into spikes. To facilitate the use of traditional sensors, we need to convert the output signals into streams of spikes, i.e., a series of events ($+1$, $-1$) along with their corresponding timestamps. In this paper, we propose a review of the coding algorithms from a robotics perspective and further supported by a benchmark to assess their performance. We also introduce a ROS (Robot Operating System) toolbox to encode and decode input signals coming from any type of sensor available on a robot. This initiative is meant to stimulate and facilitate robotic integration of neuromorphic AI, with the opportunity to adapt traditional off-the-shelf sensors to spiking neural nets within one of the most powerful robotic tools, ROS. 
\end{abstract}

\section{Introduction}

Neuromorphic artificial intelligence (AI) is the so-called third generation of AI, and it yields a wide range of incredible opportunities for solving technical and technological challenges, particularly in robotics and autonomous systems. The current second generation of AI has been focusing on the applications of deep-learning networks to analyse large datasets, with outstanding applications to sensing and perception. However, the high performances achieved by the proposed solutions come with major flaws for robotics applications. First and foremost, the computational needs of these solutions is extremely high, especially for visual-based deep networks, and thus online learning is, in most cases, prohibited. Hardware solutions have been proposed but they remain often expensive and/or beyond the robots' affordable payload (e.g., micro air vehicles). Moreover, as artificial neural networks (ANNs) are executed in a fully synchronized manner, most of the computation is inherently uselessly done. For instance, a visual-based ANN performing object tracking analyses the entire image at each timestamp while most of the information remains unchanged within small time intervals. Another consequence of the synchronous execution is that traditional ANNs are not optimized to extract the timing information. In contrast, neuromorphic systems are characterized by the asynchronous and independent execution of the neurons within the neural net, therefore enabling more flexibility, as well as the ability to learn the timing information. In this respect, spiking neural networks (SNNs) feature biologically inspired neurons that fire independently and asynchronously of the others. So far, SNNs have been widely studied in simulations, thanks to the booming efforts in designing convenient simulation environments like the Python-based APIs NORSE~\cite{norse2021}, Brian~\cite{goodman2008brian}, Nengo~\cite{bekolay2014nengo}, or PySNN\footnote{\url{https://github.com/BasBuller/PySNN}}. Unfortunately, the performances of SNNs remain limited for robotic applications, the major reason being that SNNs are executed on conventional, synchronous hardware. 

Over the past few years, intense efforts have been put in the design of neuromorphic hardware, such as HICANN~\cite{schemmel2010wafer}, NeuroGrid~\cite{benjamin2014neurogrid}, TrueNorth~\cite{merolla2014million}, SpiNNaker~\cite{furber2014spinnaker}, Loihi~\cite{davies2018loihi}, and SPOON~\cite{frenkel202028}. Although these neuromorphic chips remain under development, some of them are getting integrated onboard robots to push the frontiers of neuromorphic sensing and control in robotics~\cite{stagsted2020event, dupeyroux2020neuromorphic, michaelis2020robust, polykretis2020astrocyte}. Early results show outstanding performances, such as extremely fast execution (up to several hundred kHz) along with high energy efficiency. Yet, sensing may represent the ultimate bottleneck of neuromorphic AI in robotics. As a matter of fact, only a very limited number of neuromorphic sensors is available, the vast majority of which being dedicated to vision only (i.e., event-based cameras~\cite{9138762}). The current lack of technology for neuromorphic sensing in robotics (IMU, sonar, radar, Lidar, etc.) can be tackled in a quite efficient way by means of spike coding algorithms~\cite{petro2019selection, meftah2013image, liu2016benchmarking}, which allow to convert traditional sensors data into a stream of spikes, that is, a series of events determined by their timestamp and their polarity ($+1$ or $-1$). Although the overall performances will be hampered by the sensors' sampling frequency, it is worth noting that spike encoding, and decoding algorithms offer the opportunity to investigate novel neuromorphic algorithms while benefiting from standard off-the-shelf sensors. 

Spike coding algorithms can be divided into three categories. Population coding is the most straight-forward approach. In population coding, a set of distinct neurons is defined to encode (or decode) the signal. Each of these neurons is characterized by a specific distribution of response given the input signal. As a result, at each timestamp, only the neuron giving the maximal response to the current value of the signal will emit a spike. The remaining categories are based on the temporal variations of the signal. First, temporal coding algorithms allow to encode the information with high timing precision. The core idea of temporal coding is that spikes will be emitted whenever the variation of the signal gets higher than a threshold. Lastly, rate coding is used to encode the information into the spiking rate. An alternative to these algorithms is to directly setup the encoding and decoding layers in the simulated SNN, and then train/evolve the network itself. However, this approach is case-specific and may not be supported by neuromorphic hardware.

In this paper, we propose an overview of the different neuromorphic coding solutions available for applications in robotics, with a set of MATLAB scripts and Python codes implementing both encoding and decoding algorithms. A benchmark is proposed to assess the interests of each of the implemented methods, focusing on crucial aspects in field robotics like how the information is encoded, but also the computational requirements and the coding efficiency (i.e., quality of the signal reconstruction, number of spikes generated, etc.). Lastly, we introduce a ROS (Robot Operating System) package implementing these algorithms to enhance the use of neuromorphic solutions for robotics with traditional sensors (see \hyperref[sec:supplementary]{Supplementary Materials}).

\section{Spike coding schemes overview}

    \subsection{Population coding schemes}

Population coding has been shown to be widely used in the brain (sensor, motor)~\cite{georgopoulos1986neuronal, averbeck2006neural}. It suggests that neurons (or populations of neurons) activity is determined by a distribution over the input signal. A typical example of such coding has been extensively studied in ganglion cells of the retina~\cite{nirenberg1998population, warland1997decoding}, known to be the only neurons involved in the transmission of the visual information from the retina to the brain. Studies in the somatosensory cortex, responsible for processing somatic sensations, also revealed the existence of population coding~\cite{petersen2002population}. 

In this work, we implemented a two simple models of population coding, position-coding and Gaussian Receptive Fields, further referred to as GRF. 

\begin{enumerate}
    \item[(i)] Position coding is the simplest form of population coding. In this encoding method, a group of neurons all get assigned a small part of the underlying distribution (Algorithm~\ref{alg:pos}). This distribution can be linearly divided but also non-linear distributions are possible. While having a linear distribution allows to encode a signal in a homogeneous way, one could consider implementing a Gaussian distribution instead. Therefore, the encoding would be better suited for neuromorphic control, with input being the error to a targeted set point (e.g., thrust control based on the divergence of the optic flow field in a drone landing task~\cite{dupeyroux2020neuromorphic}).  
    \item[(ii)] Inspired by the aforementioned studies, the GRF model encodes the signal by means of a set of neurons which activity distributions are defined as Gaussian waves determined by a centre $\mu$ and a variance $\sigma^2$. In the proposed model, all centres $\mu_i$ of neurons $i$ are regularly spaced to cover the maximum amplitude of the input signal, while the variances are set equal. The sequence between the regular timesteps is now split in a number of sub-timesteps that allow for a combination of position- and temporal encoding. Neurons with a high activation will fire at the beginning, and neurons with a lower activation later (or not at all if the response is below a certain threshold) (Algorithm~\ref{alg:grf}). If a temporal component is not desirable or possible for the application, it can also be chosen to map the neurons activation value to a probability of spiking, meaning that neurons closer to the encoded value have a larger probability of spiking. 
\end{enumerate}
 
     \begin{algorithm}[!t]
    \SetAlgoLined
    \DontPrintSemicolon
    \footnotesize
    \caption{Position-coding}
    \label{alg:pos}
     \KwData{$input$, $distribution$}
     \KwResult{$spikes$, $minVal$, $maxVal$}
     $L \gets length(input)$\;
     $spikes \gets zeros(L,m)$\;
     $distances \gets zeros(m)$\;
     \For{$j = 1:L$}{
        \For{$i = 1:length(distribution)$}{
            $distances(i) = sqrt((input(j) - distribution(i))^2)$\;
        }
        $[\_,~spikingNeuron] = min(distances)$\;
        $spikes(j,spikingNeuron) = 1$\;
     }
    \end{algorithm} 
 
    \begin{algorithm}[!t]
    \SetAlgoLined
    \DontPrintSemicolon
    \footnotesize
    \caption{GRF-based population spike encoding}
    \label{alg:grf}
     \KwData{$input$, $m$, $n$}
     \KwResult{$spikes$, $minSig$, $maxSig$}
     $L \gets length(input)$\;
     $[minSig,~maxSig] \gets [min(input),~max(input)]$\;
     $spikes \gets zeros(L,n,m)$\;
     $responses \gets zeros(m)$\;
     \For{$i = 1:m$}{
        $\mu(i) = minSig + (2\cdot (i+1)-3)/2\cdot (maxSig - minSig)/(m-2)$\;
        $\sigma(i) = (maxSig - minSig)/(m-2)$\;
     }
     $maxProb \gets norm(\mu(0), \mu(0), sigma(0)$\;
     $timingDistribution \gets linspace(0, maxProb, n + 1)$\;
     \For{$j = 1:L$}{
        \For{$i = 1:m$}{
            $responses(i) = norm(input(j), \mu(i), \sigma(i))$\;
            $distances(i) = sqrt((responses(i) - timingDistribution(i))^2)$\;
            $spikeTime(i) = n - 1 - min(distances(i)$\;
            \If{$spikeTime(i) < n - 1$}{
                $spikes(j, spikeTime(i), i) = 1$ \;
            }
        }
     }
    \end{algorithm} 

    \subsection{Rate coding schemes}

According to the rate coding paradigm, the information is encoded in the spiking rate instead of the specific spike timing. In this paper, we will focus on the following coding algorithms:\begin{enumerate}
    \item[(i)] The Hough Spike Algorithm (HSA): equivalent to a threshold-based mode, the HSA makes use of a reverse convolution between the buffered input signal with a finite impulse response (FIR) filter to determine the spikes timings (Algorithm~\ref{alg:hsa})~\cite{hough1999spiker}.
    \item[(ii)] The Threshold Hough Spike Algorithm (T-HSA): similar to the HSA, the T-HSA introduces a threshold to compare with the error between the signal and the filter (Algorithm~\ref{alg:t-hsa})~\cite{schrauwen2003bsa}. Whenever the error exceeds this threshold, a spike is emitted, and the input signal is updated by subtracting the filter response. In the T-HSA, the threshold depends on the signal and must be determined prior to the encoding.
    \item[(iii)] The Ben's Spike Algorithm (BSA): as for the previous coding schemes, the BSA applies a reverse convolution of the signal with a FIR filter. To determine when a spike must be generated, the algorithm uses two errors, the first one being the sum of differences between the signal and the filter, and the second one being the sum of the signal values. The algorithm then generates spikes by comparing the first error to a fraction of the accumulated signal, defined as the product between the second error and a predefined threshold (Algorithm~\ref{alg:bsa})~\cite{schrauwen2003bsa}. Unlike T-HSA, the threshold is filter-dependent, allowing to keep the same value for different signals.
\end{enumerate}

    \begin{algorithm}[!t]
    \SetAlgoLined
    \DontPrintSemicolon
    \footnotesize
    \caption{Hough spike encoding}
    \label{alg:hsa}
     \KwData{$input$, $filter$}
     \KwResult{$spikes$, $shift$}
     $L \gets length(input)$\;
     $F \gets length(filter)$\;
     $spikes \gets zeros(1,L)$\;
     $shift \gets min(input)$\;
     $input \gets input - shift$\;
     \For{$i = 1:L$}{
        $count \gets 0$\;
        \For{$j = 1:F$}{
            \If{$i+j+1 \leq L$ \& $input(i+j-1) \geq filter(j)$}{
                $count += 1$\;
            }
            \If{$count == F$}{
                $spikes(i) = 1$\;
                \For{$j = 1:F$}{
                    \If{$i+j+1 \leq L$}{
                        $input(i+j-1) = input(i+j-1) - filter(j)$\;
                    }
                }
            }
        }
     }
    \end{algorithm}
    
    \begin{algorithm}[!t]
    \SetAlgoLined
    \DontPrintSemicolon
    \footnotesize
    \caption{Threshold Hough spike encoding}
    \label{alg:t-hsa}
     \KwData{$input$, $filter$, $threshold$}
     \KwResult{$spikes$, $shift$}
     $L \gets length(input)$\;
     $F \gets length(filter)$\;
     $spikes \gets zeros(1,L)$\;
     $shift \gets min(input)$\;
     $input \gets input - shift$\;
     \For{$i = 1:L$}{
        $error \gets 0$\;
        \For{$j = 1:F$}{
            \If{$i+j+1 \leq L$ \& $input(i+j-1) \geq filter(j)$}{
                $error += filter(j) - input(i+j-1)$\;
            }
            \If{$error \leq threshold$}{
                $spikes(i) = 1$\;
                \For{$j = 1:F$}{
                    \If{$i+j+1 \leq L$}{
                        $input(i+j-1) = input(i+j-1) - filter(j)$\;
                    }
                }
            }
        }
     }
    \end{algorithm}
    
    \begin{algorithm}[!t]
    \SetAlgoLined
    \DontPrintSemicolon
    \footnotesize
    \caption{Ben's spike encoding}
    \label{alg:bsa}
     \KwData{$input$, $filter$, $threshold$}
     \KwResult{$spikes$, $shift$}
     $L \gets length(input)$\;
     $F \gets length(filter)$\;
     $spikes \gets zeros(1,L)$\;
     $shift \gets min(input)$\;
     $input \gets input - shift$\;
     \For{$i = 1:(L-F)$}{
        $err1,~err2 \gets 0$\;
        \For{$j = 1:F$}{
            $err1 = err1 + abs(input(i+j) - filter(j))$\;
            $err2 = err2 + abs(input(i+j-1))$\;
        }
        \If{$err1 \leq err2 \cdot threshold$}{
           $spikes(i) = 1$\;
           \For{$j=1:F$}{
                \If{$i+j+1 \leq L$}{
                    $input(i+j+1) = input(i+j+1) - filter(j)$\;
                }
           }
        }
     }
    \end{algorithm}

While rate coding is considered to be the universal way neurons encode the information, many studies highlighted the poor performances of rate coding schemes as compared to temporal coding algorithms~\cite{gautrais1998rate, rullen2001rate}. 
   
    \subsection{Temporal coding schemes}

Temporal representations, also called pulse coding, provide a time-based coding where the information is encoded in the exact spikes timing. Unlike rate coding, temporal coding provides more information capacity~\cite{abeles1994synchronization}, and is further supported by neuro-physiological studies showing that auditory and visual information are processed with high precision in the brain~\cite{carr1993processing}.

Several models have been proposed. Assuming that the most significant information is carried by the first spikes, the Rank Order Coding (ROC) arranges spikes with respect to their arrival time~\cite{rullen2001rate, perrinet2004coding}. Taking inspiration from the ganglion cells, the Latency-Phase Coding (LPC) was introduced to combine the exact time spiking provided by temporal coding with the phase information (encoding spatial information in the ganglion cells)~\cite{nadasdy2009information, hu2013spike}. Phase encoding has also been implemented in~\cite{wu2018spiking} to investigate on neurons in the human auditory cortex. In this paper, we will focus on the following coding algorithms:\begin{enumerate}
    \item[(i)] The Temporal-Based Representation (TBR): also called temporal contrast, the TBR algorithm generates spikes whenever a the variation of the signal, between two consecutive timestamps, gets higher than a fixed threshold (Algorithm~\ref{alg:temporal_contrast})~\cite{lichtsteiner200564x64}. Event-based cameras such as the Dynamic Vision Sensor (DVS) implement this coding scheme to generate a stream of events at extremely fast speed~\cite{lichtsteiner200564x64, lichtsteiner2008128}.
    \item[(ii)] The Step Forward Algorithm (SF): based on the TBR coding scheme, the SF uses a baseline signal to compute the variation of the input signal (Algorithm~\ref{alg:step_forward})~\cite{kasabov2016evolving}. As for the TBR model, a spike ($+1$ or $-1$) is emitted whenever the variation exceeds the threshold. Simultaneously, the baseline gets updated by $\pm$ the threshold depending on the spike polarity. 
    \item[(iii)] The Moving Window Algorithm (MW): the MW model is similar to the SF model, but here the baseline signal is defined as the mean of the previous signal intensities over a time window (Algorithm~\ref{alg:moving_window})~\cite{kasabov2016evolving}. 
\end{enumerate}

    \begin{algorithm}[!t]
    \SetAlgoLined
    \DontPrintSemicolon
    \footnotesize
    \caption{Temporal contrast encoding}
    \label{alg:temporal_contrast}
     \KwData{$input$, $\alpha$}
     \KwResult{$spikes$, $threshold$}
     $L \gets length(input)$\;
     $spikes \gets zeros(1,L)$\;
     $diff \gets zeros(1,L-1)$\;
     \For{$i = 1:L-1$}{
        $diff(i) = input(i+1) - input(i)$\;
     }
     $threshold \gets mean(diff) + \alpha \cdot std(diff)$\;
     $diff \gets [diff(1),~diff]$\;
     \For{$i = 1:L-1$}{
        \uIf{$diff(i) > threshold$}{
           $spikes(i) = 1$\;
        }\uElseIf{$diff(i) < -threshold$}{
           $spikes(i) = -1$\;
        }
     }
    \end{algorithm}
    
    \begin{algorithm}[!t]
    \SetAlgoLined
    \DontPrintSemicolon
    \footnotesize
    \caption{Step-forward encoding}
    \label{alg:step_forward}
     \KwData{$input$, $threshold$}
     \KwResult{$spikes$, $init$}
     $L \gets length(input)$\;
     $spikes \gets zeros(1,L)$\;
     $init,~base \gets input(1)$\;
     \For{$i = 2:L$}{
        \uIf{$input(i) > base + threshold$}{
           $spikes(i) = 1$\;
           $base = base + threshold$\;
        }\uElseIf{$input(i) < base - threshold$}{
           $spikes(i) = -1$\;
           $base = base + threshold$\;
        }
     }
    \end{algorithm}

    \begin{algorithm}[!t]
    \SetAlgoLined
    \DontPrintSemicolon
    \footnotesize
    \caption{Moving window encoding}
    \label{alg:moving_window}
     \KwData{$input$, $window$, $threshold$}
     \KwResult{$spikes$, $init$}
     $L \gets length(input)$\;
     $spikes \gets zeros(1,L)$\;
     $init \gets input(1)$\;
     $base \gets mean(input(1:(window+1)))$\;
     \For{$i = 1:window+1$}{
        \uIf{$input(i) > base + threshold$}{
           $spikes(i) = 1$\;
        }\uElseIf{$input(i) < base - threshold$}{
           $spikes(i) = -1$\;
        }
     }
     \For{$i = window+2:L$}{
        $base = mean(input((i-window-1):(i-1)))$\;
        \uIf{$input(i) > base + threshold$}{
           $spikes(i) = 1$\;
        }\uElseIf{$input(i) < base - threshold$}{
           $spikes(i) = -1$\;
        }
     }
    \end{algorithm}

Since SF and MW models feature an adaptive component in the calculation of the signal variations by means of the baseline, they are known to result in better reconstruction of the encoded signal after decoding. 

    \subsection{Comparison of the selected schemes}

\begin{figure*}
    \centering
    \includegraphics[width=\textwidth]{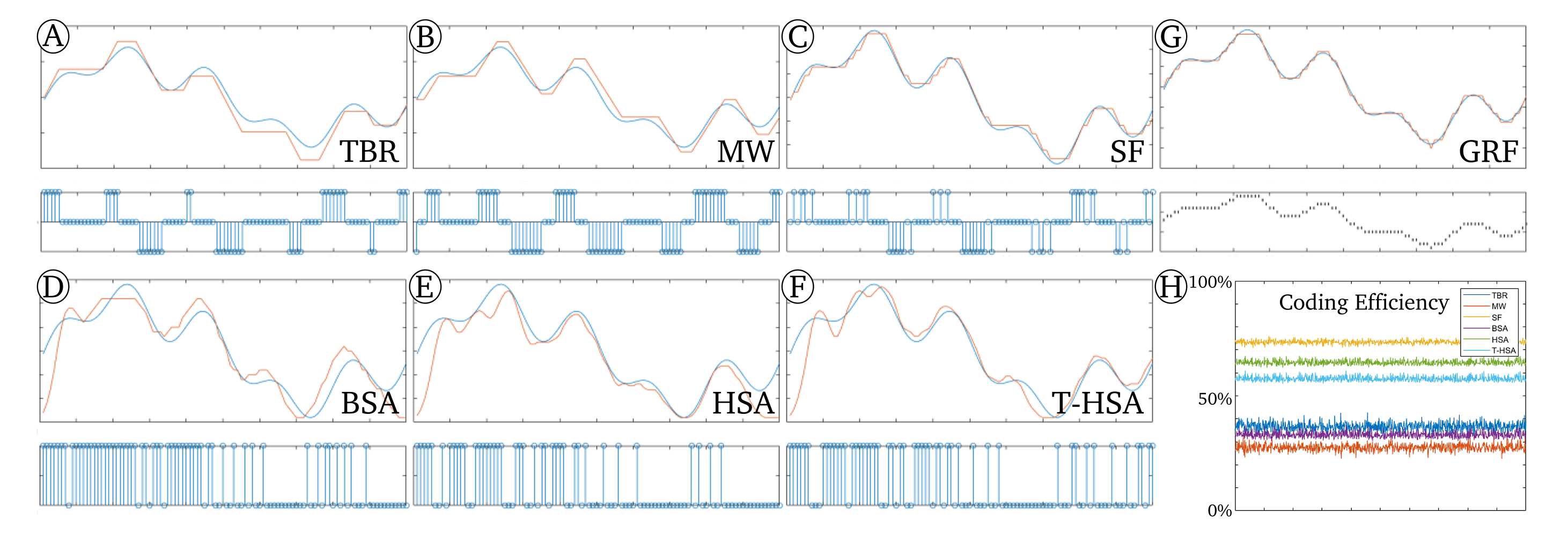}
    \caption{Example of the typical outputs of the selected coding schemes on a 1D signal. \textbf{A-C} Temporal coding. \textbf{D-F} Rate coding. \textbf{G} Population coding. For each coding scheme (\textbf{A-G}), the original (\textit{blue}) and the reconstructed (\textit{orange}) signals are displayed (\textbf{top}), as well as the corresponding spikes (\textbf{bottom}). \textbf{H} Coding efficiency of the different methods, expressed as a percentage. \textit{Yellow}: SF. \textit{Green}: HSA. \textit{Cyan}: T-HSA. \textit{Blue}: TBR. \textit{Purple}: BSA. \textit{Orange}: MW.}
    \label{fig:sample_tests}
\end{figure*}

The selected algorithms have been implemented both in MATLAB and Python 3 (encoding and decoding algorithms; see \hyperref[sec:supplementary]{Supplementary Materials}). In this section, we propose to investigate on their performances by means of a benchmark over a set of 1D signals (sum of noisy sine waves). An overview of the typical outputs provided by each algorithm is shown in Fig.~\ref{fig:sample_tests}(A-G). Qualitatively, it is worth noting that both the quality of the signal reconstruction and the efficiency in spike encoding vary from one model to another. In the following, we define spiking efficiency as the percentage of timestamps without spike emission: 

\begin{equation}
    \text{spiking efficiency} = \Big( 1 - \frac{spike~count}{length~of~the~signal}\Big) \times 100
\end{equation}

These observations are further investigated by assessing the spiking efficiency of each algorithm as well as the root mean squared error (RMSE) between the input signal and the reconstructed one over a set of $N = 1000$ samples. The long-term drift, due to cumulative errors in the signal reconstruction, are also considered by extending the duration of the signal from 5~seconds to 100~seconds. Statistical results are given in Tables~\ref{table:spike_efficiency} (spiking efficiency) and~\ref{table:rmse} (RMSE). First, we observe that algorithms SF, HSA, and T-HSA result in a high spiking efficiency ($>50\%$). In particular, we note that the SF model is not affected by the duration of the signal (stable at $73\%$), while the spiking efficiency of rate coding HSA and T-HSA drops 13 (HSA) and 7 (T-HSA) points, respectively. In contrast, the MW scheme maintains a spiking efficiency of $27\%$, regardless of the signal duration. Lastly, it is interesting to note that the standard deviation of the spiking efficiency tends to 0 when the signal duration gets bigger. This suggest that for each algorithm, and certainly for each \textit{type} of signal, an optimal duration of the signal ensures stable spiking efficiency. Fig~\ref{fig:sample_tests}(H) shows an example of the spiking efficiency over the 1000 sample tests (case 3). 

\begin{table}[!h]
\centering
\caption{Average spiking efficiency of the coding schemes \\w.r.t. the signal duration}
\label{table:spike_efficiency}
\begin{tabular}{lcccccc}
 & \textbf{TBR} & \textbf{MW} & \textbf{SF} & \textbf{BSA} & \textbf{HSA} & \textbf{T-HSA} \\ \hline
\multicolumn{7}{l}{\textit{Case \#1 -- T\_max = 1 sec (1 period), nb\_tests = 1000}} \\
\textit{Mean} & 36.7 & 29.2 & 74.6 & 45.8 & 74.0 & 62.4 \\
\textit{SD} & 4.1 & 3.4 & 2.0 & 1.8 & 1.3 & 1.1 \\ \hline
\multicolumn{7}{l}{\textit{Case \#2 -- T\_max = 5 sec (5 periods), nb\_tests = 1000}} \\
\textit{Mean} & 36.7 & 27.5 & 73.4 & 33.1 & 64.5 & 57.5 \\
\textit{SD} & 1.7 & 1.5 & 0.9 & 1.3 & 1.0 & 1.1 \\ \hline
\multicolumn{7}{l}{\textit{Case \#3 -- T\_max = 15 sec (15 periods), nb\_tests = 1000}} \\
\textit{Mean} & 36.6 & 27.2 & 73.2 & 30.0 & 62.1 & 55.8 \\
\textit{SD} & 1.0 & 0.9 & 0.5 & 0.8 & 0.6 & 0.6 \\ \hline
\multicolumn{7}{l}{\textit{Case \#4 -- T\_max = 50 sec (50 periods), nb\_tests = 1000}} \\
\textit{Mean} & 36.7 & 27.1 & 73.2 & 29.3 & 61.6 & 55.6 \\
\textit{SD} & 0.6 & 0.5 & 0.3 & 0.4 & 0.3 & 0.3 \\ \hline
\multicolumn{7}{l}{\textit{Case \#5 -- T\_max = 100 sec (100 periods), nb\_tests = 1000}} \\
\textit{Mean} & 36.6 & 27.1 & 73.2 & 29.0 & 61.4 & 55.5 \\
\textit{SD} & 0.4 & 0.3 & 0.2 & 0.2 & 0.2 & 0.2 \\ \hline
\end{tabular}
\end{table}

\begin{table}[!h]
\centering
\caption{Average RMSE between the original and the \\reconstructed signals w.r.t. the signal duration}
\label{table:rmse}
\begin{tabular}{lcccccc}
 & \textbf{TBR} & \textbf{MW} & \textbf{SF} & \textbf{BSA} & \textbf{HSA} & \textbf{T-HSA} \\ \hline
\multicolumn{7}{l}{\textit{Case \#1 -- T\_max = 1 sec (1 period), nb\_tests = 1000}} \\
\textit{Mean} & 0.80 & 0.68 & 0.26 & 0.84 & 0.97 & 0.68 \\
\textit{SD} & 0.39 & 0.24 & 0.01 & 0.04 & 0.05 & 0.03 \\ \hline
\multicolumn{7}{l}{\textit{Case \#2 -- T\_max = 5 sec (5 periods), nb\_tests = 1000}} \\
\textit{Mean} & 1.41 & 1.11 & 0.26 & 0.64 & 0.62 & 0.39 \\
\textit{SD} & 0.68 & 0.54 & 0.01 & 0.02 & 0.03 & 0.01 \\ \hline
\multicolumn{7}{l}{\textit{Case \#3 -- T\_max = 15 sec (15 periods), nb\_tests = 1000}} \\
\textit{Mean} & 2.24 & 1.74 & 0.26 & 0.60 & 0.51 & 0.30 \\
\textit{SD} & 1.17 & 0.86 & 0.00 & 0.01 & 0.02 & 0.06 \\ \hline
\multicolumn{7}{l}{\textit{Case \#4 -- T\_max = 50 sec (50 periods), nb\_tests = 1000}} \\
\textit{Mean} & 3.90 & 2.94 & 0.26 & 0.58 & 0.48 & 0.26 \\
\textit{SD} & 2.00 & 1.57 & 0.00 & 0.00 & 0.01 & 0.00 \\ \hline
\multicolumn{7}{l}{\textit{Case \#5 -- T\_max = 100 sec (100 periods), nb\_tests = 1000}} \\
\textit{Mean} & 5.55 & 4.14 & 0.26 & 0.57 & 0.46 & 0.25 \\
\textit{SD} & 2.98 & 2.17 & 0.00 & 0.00 & 0.01 & 0.00 \\ \hline
\end{tabular}
\end{table}

\begin{figure}
    \centering
    \includegraphics[width=0.8\linewidth]{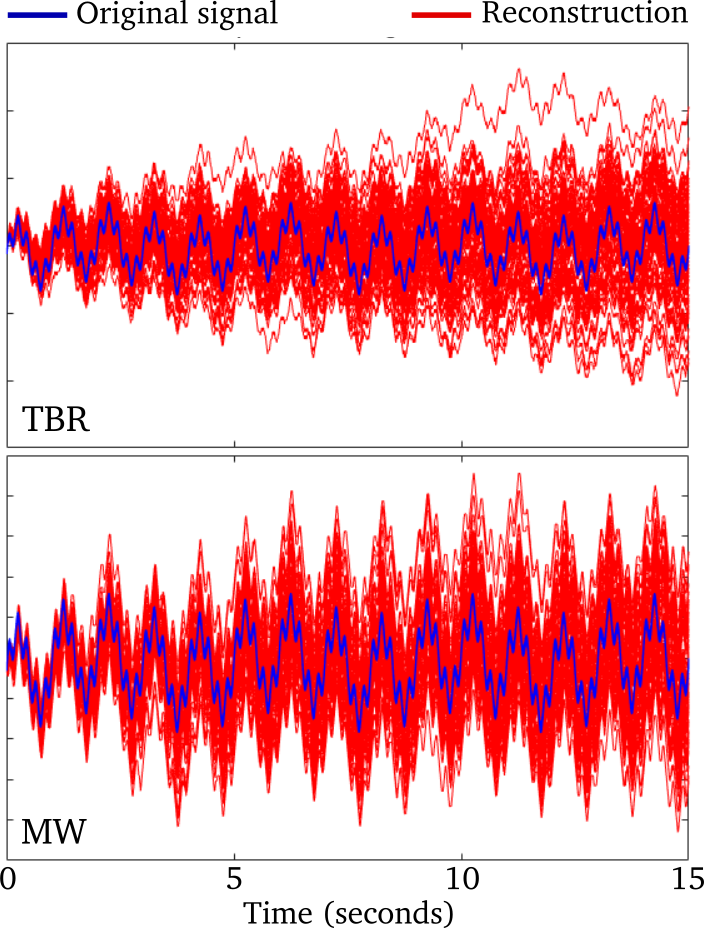}
    \caption{Typical drift of the reconstructed signals for both the TBR (\textbf{top}) and the MW (\textbf{bottom}) temporal coding schemes ($N=1000$ samples).}
    \label{fig:drift}
\end{figure}

The quality of the reconstruction is reflected by the RMSE. Once again, the SF model demonstrates the best performances, showing no significant effect of the signal duration, and with an overall RMSE as low as $0.26 \pm 0.01$ (mean $\pm$ standard deviation), while the amplitude of the input signal is equal to $5.8$. Besides, we note that the performance of the rate coding (BSA, HSA, and T-HSA) tend to improve as the signal duration increases, with a final average RMSE inferior to $0.6$. As for the spiking efficiency, algorithms SF, BSA, HSA, and T-HSA tend to stabilize their mean RMSE with increased signal duration, thus reflecting the existence of an optimum. However, the temporal coding TBR and MW show an increasing RMSE, reaching the amplitude of the input signal itself after 100~periods. As shown in Fig.~\ref{fig:drift}, these algorithms accumulate errors over time and tend to drift from the input, while the wave form of the signal is maintained. 

Lastly, the GRF population coding scheme shows accurate signal reconstruction for as long as enough encoding neurons are available (Fig.~\ref{fig:sample_tests}G). However, the computational cost grows with the coding neurons, and the spiking efficiency is null: a spike is emitted at each timestamp, resulting in a loss in the spike timing accuracy. 

    \subsection{Summary}

When it comes to robotic applications, it is of high importance to balance the overall performances of the coding schemes with the available computational resources and the type of information to process. For instance, the GRF population coding algorithm allows to precisely encode the value of the input signal, while temporal and rate coding schemes will rather encode the temporal variations. Rate coding schemes can be greedy in the way that they require buffered signals to achieve good performances. They also suffer from major limitations as they are inefficient (i.e., they produce a great number of spikes), and their performances depend on the type of filter (which itself depends on the data encoded). On the positive side, they are quite robust to disturbances (very low RMSE after reconstruction). Table~\ref{table:params} provides an overview of the parameters required to deploy each algorithm. These parameters care mostly case specific and must be determined prior to the online robotic application. A straightforward approach for this would be the use of evolutionary algorithms within Python-based frameworks like DEAP~\cite{fortin2012deap}, PyBrain~\cite{schaul2010pybrain} and PyEvolve~\cite{perone2009pyevolve}.

\begin{table}[!h]
\centering
\caption{Parameters required for each coding scheme}
\label{table:params}
\begin{tabular}{lccccccc}
 & \textbf{TBR} & \textbf{MW} & \textbf{SF} & \textbf{BSA} & \textbf{HSA} & \textbf{T-HSA} & \textbf{GRF} \\ \hline
\textit{Factor} & x &  &  &  &  &  &  \\
\textit{Threshold} &  & x & x & x &  & x &  \\
\textit{Window} &  & x &  &  &  &  &  \\
\textit{Filter} &  &  &  & x & x & x &  \\
\textit{Neurons} &  &  &  &  &  &  & x \\ \hline
\end{tabular}
\end{table}

\section{A ROS package proposal}

    \subsection{Description of the package}
    
With the aim of reducing the sensing bottleneck and therefore stimulating investigations in neuromorphic AI for robotics, we propose an ROS implementation of the aforementioned coding schemes (\textit{GRF}, \textit{TBR}, \textit{MW}, \textit{SF}, \textit{BSA}, \textit{HSA}, and \textit{T-HSA}; see \hyperref[sec:supplementary]{Supplementary Materials}). Both encoding and decoding algorithms are implemented in Python, in the same scripts as those described in the previous section. The provided ROS tools are organized as follows:\begin{enumerate}
    \item[(i)] The \textit{spyke\_msgs} package introduces two new type of ROS messages, i.e. \textit{spyke.msg} which contains the spike ($+1$, $0$, $-1$), the corresponding timestamp (in seconds), and a set of parameters that depend on the signal and the coding scheme. The second message, \textit{spyke\_array.msg} is a copy of the previous one but designed for carrying an array of spikes instead.
    \item[(ii)] The \textit{spyke\_coding\_schemes} package contains the encoding and decoding functions to be installed in the ROS workspace. 
    \item[(iii)] The \textit{spyke\_coding} package defines the ROS node to be launched to start encoding or decoding an input signal. Once active, the ROS node will publish the message and record the data in a rosbag located at $\mathtt{\sim}$\textit{/.ros/}. We also provide a Python script to process the data.
\end{enumerate}

The ROS node can be launched as usual by running the following command: \textit{roslaunch~spyke\_coding~$<$launchfile$>$}. A set of launch files for each implemented coding scheme is available within the \textit{spyke\_coding} package. To facilitate the use of the package, the ROS node (\textit{spyke\_coding/src/generate\_spikes.py}) simulates a 1D signal to be encoded. This can be easily replaced by a ROS subscriber to any sensor available.

    \subsection{Examples}

\begin{figure}
    \centering
    \includegraphics[width=0.85\linewidth]{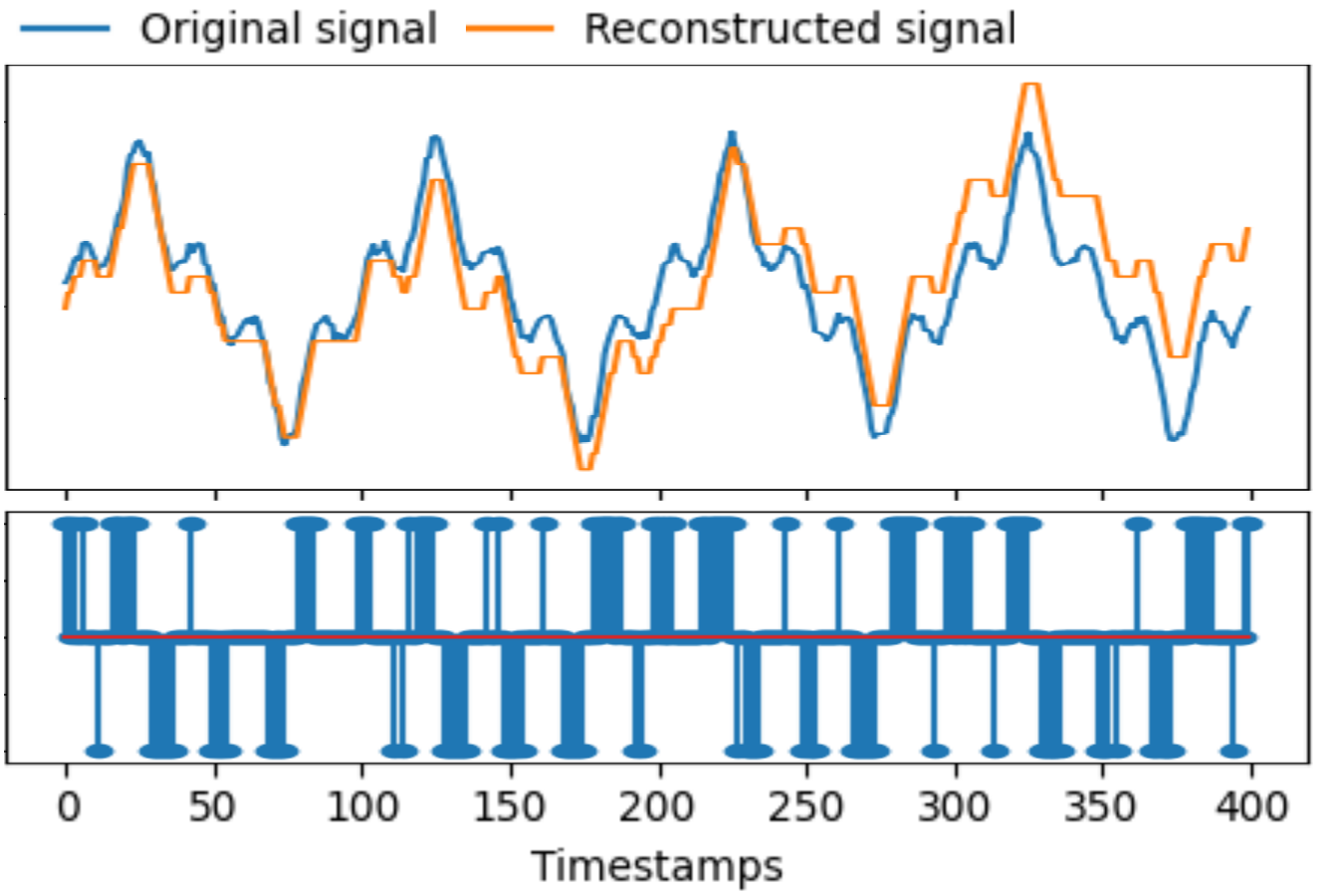}
    \caption{Example of the spike generation in ROS using the SF algorithm. \textbf{(Top)} Graphic representation of the original (blue) and reconstructed (orange) signals. \textbf{(Bottom)} Spikes generated by the encoding scheme. Time: one timestamp equals 0.01 seconds.}
    \label{fig:ros_sf}
\end{figure}

Here we provide an example of the output signals provided by the ROS toolbox. In this example, we consider an input signal of 4~seconds with a sampling frequency of 100~Hz. The signal is defined as the sum of three sine waves of frequencies 1~Hz, 2~Hz, and 5~Hz. Noise is added to the signal. The encoding scheme selected is the SF model for which the threshold is equal to 0.35. A Python script is available for automatic processing of rosbags. In Fig.~\ref{fig:ros_sf}, the recorded data, i.e., the input signal and the generated spikes, are displayed along with the reconstructed signal.

\begin{figure}
    \centering
    \includegraphics[width=\linewidth]{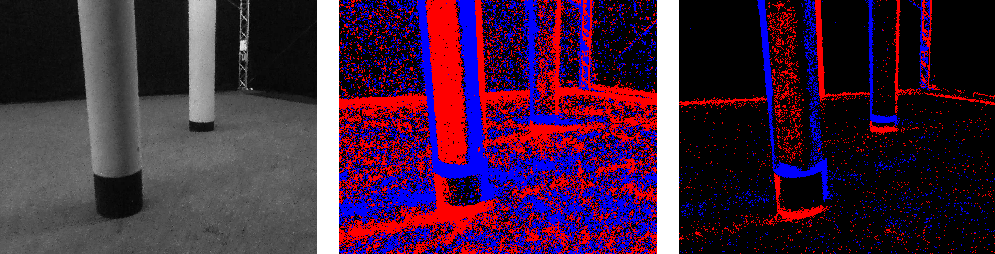}
    \caption{Example of the application of the SF algorithm to a video stream. \textbf{(Left)} Original video input. \textbf{(Middle)} Encoding with a threshold of 1. \textbf{(Right)} Encoding with a threshold of 10.}
    \label{fig:video}
\end{figure}

In Fig.~\ref{fig:video}, we provide an example of the effect of the use of the SF coding scheme on a video input for varying threshold. The video is part of the open-source Obstacle and Avoidance (ODA) Dataset (\url{https://github.com/tudelft/ODA_Dataset}). The toolbox can therefore help emulating the so-called event-based cameras and ensures an easy tuning of the encoding of the visual input with respect to both the robotic application and the environmental conditions. 

\section{Conclusions and future work}

We introduced a toolbox for neuromorphic coding for sensing in robotics with the aim to facilitate the development of fully neuromorphic systems onboard robots, from perception to action. It includes the following algorithms: the position coding, the Gaussian Receptive Fields (GRF) population coding, the Temporal-Based Representation (TBR, also used in event-based cameras), the Step-Forward (SF) and the Moving Window (MW) algorithms, as well as the following rate-based coding schemes: the Ben's Spike Algorithm (BSA), the Hough Spike Algorithm (HSA) and the Threshold Hough Spike Algorithm (T-HSA). The toolbox contains implementations of the encoding and decoding algorithms in MATLAB and Python, and has been integrated to the ROS framework to encode and decode signals online onboard robots using off-the-shelf sensors. A benchmark was proposed to assess the advantages and drawbacks of each of the proposed coding schemes. This toolbox is expected to evolve, with the will to integrate other coding schemes like the non-linear GRF population coding, the Rank Order Coding (ROC), the Latency-Phase temporal Coding (LPC). 

\section*{Supplementary materials}
\label{sec:supplementary}
The MATLAB, Python and ROS codes are available at: \url{https://github.com/tudelft/SpikeCoding}.

\bibliographystyle{IEEEtran}
\bibliography{myRefs}

\end{document}